\documentclass[letterpaper, 10 pt, journal, twoside]{ieeetran}

\author{Davide Nardi$^{1*}$, Edoardo Lamon$^{1*}$, Daniele Fontanelli$^{2}$, Matteo Saveriano$^{2}$, and Luigi Palopoli$^{1}$
\thanks{Manuscript received: June, 30, 2025; Revised October, 15, 2025; Accepted December, 18, 2025.
This paper was recommended for publication by Editor J. Burgner-Kahrs upon evaluation of the Associate Editor and Reviewers’ comments.
This work was supported by the MUR PNRR project FAIR - Future AI Research (PE00000013), the project iNEST - Interconnected Nord-Est Innovation Ecosystem (ECS 00000043) funded by the NextGenerationEU, and the project INVERSE (Grant Agreement n. 101136067).}
\thanks{All the authors are with the Interdepartmental Robotics Labs (IDRA),  Universit\`a di Trento. $^{*}$ Contributed equally to this work.}
\thanks{$^{1}$Department of Information Engineering and Computer Science, Universit\`a di Trento, Trento, Italy. \tt\small davide.nardi-1@unitn.it}
\thanks{$^{2}$Department of Industrial Engineering, Universit\`a di Trento, Trento, Italy.}
\thanks{Digital Object Identifier (DOI): see top of this page.}
}
\usepackage[normalem]{ulem}

\usepackage{cite}
\usepackage{amsmath,amssymb,amsfonts}
\usepackage{bm}
\usepackage{algorithmic}
\usepackage{graphicx}
\usepackage{textcomp}
\usepackage{xcolor}
\usepackage{balance}
\usepackage{hyperref}
\usepackage{lipsum}
\usepackage{breqn}
\usepackage{booktabs}
\usepackage{siunitx}
\usepackage{import}
\usepackage{standalone}
\usepackage{tikz}
\usetikzlibrary{calc,patterns,decorations.pathmorphing,decorations.markings,positioning,backgrounds,arrows,shapes,fit,matrix,spy,shapes.geometric}
\usepackage{pgfplots}
\pgfplotsset{width=10cm,compat=1.9}

\usepackage[inline]{enumitem}

\DeclareMathOperator*{\argmin}{arg\,min}

\newcommand{\trsp}{{^{\top}}}
\newcommand{\bs}[1]{\boldsymbol{#1}}

\title{
An Anatomy-Aware Shared Control Approach for Assisted Teleoperation of Lung Ultrasound Examinations
}

\begin{document}
\bstctlcite{IEEEexample:BSTcontrol}
\maketitle


\begin{abstract}
Although fully autonomous systems still face challenges due to patients' anatomical variability, teleoperated systems appear to be more practical in current healthcare settings. This paper presents an anatomy-aware control framework for teleoperated lung ultrasound. Leveraging biomechanically accurate 3D modelling, the system applies virtual constraints on the ultrasound probe pose and provides real-time visual feedback to assist in precise probe placement tasks. A twofold evaluation, one with 5 naïve operators on a single volunteer and the second with a single experienced operator on 6 volunteers, compared our method with a standard teleoperation baseline. The results of the first one characterised the accuracy of the anatomical model and the improved perceived performance by the naïve operators, while the second one focused on the efficiency of the system in improving probe placement and reducing procedure time compared to traditional teleoperation. The results demonstrate that the proposed framework enhances the physician's capabilities in executing remote lung ultrasound, reducing more than 20\% of execution time on 4-point acquisitions, towards faster, more objective and repeatable exams.
\end{abstract}


\section{INTRODUCTION}
\IEEEPARstart{T}{he} process of remote delivery of healthcare, known also as telehealth, has benefited in the last decade from data-driven algorithms, from the enhancement of predictive diagnoses to the personalised treatments based on real-time patient data. 
On the other hand, progress in robotics, haptics, and virtual reality has demonstrated the possibility of remote visits and surgeries, where a physician can examine a patient in a different location using robotic instruments controlled via a teleoperation interface.
In the context of lung ultrasound (LUS), researchers have focused on
enhancing both teleoperated~\cite{Wang2021ApplicationPneumonia,Ye2021Feasibility2019,Tsumura2021Tele-OperativePatients} and autonomous~\cite{Al-Zogbi2021AutonomousDiseases,Tan2023FullyTriage,Zhang2023VisualSystem} robotic ultrasound technologies based on collaborative manipulators.

Despite this, fully autonomous systems are far from widespread implementation in hospitals and healthcare facilities due to their lack of robustness and predictability caused by significant anatomical variability across individuals and medical machines~\cite{jiang2023robotic}.
Telepresence and teleoperated robotic systems, instead, have proven to be effective solutions, especially in cases where human decision-making is still superior to the level of autonomy reached by the intelligence of such systems.
However, tele-echography requires trained experts capable of not only operating remote machines but also replicating complex medical procedures. Due to these constraints, the number of experts is usually limited and below demand. For this reason, user-friendliness and the ability to provide reliable feedback~\cite{fu2023robot}, assist~\cite{huang2024robot-assisted} and train~\cite{shahbazi2013Dual-user} the operator are of paramount importance.
Virtual fixtures (VF) represent one of the widely used tools to provide assistance to a remote operator by constraining a teleoperated robot to stay within/avoid certain regions in space~\cite{bowyer2014active} or maintain a target orientation~\cite{shen2024safe}, in a shared-control fashion.
Although initial approaches envisioned the prior computations of such regions~\cite{Li2007SpatialAnatomy}, more recent ones tried to exploit RGB images~\cite{bettini2004vision,moccia2020vision-based} and point clouds~\cite{ryden2012forbidden,Kastritsi2024PassiveDelays} to initialise them. However, these methods assume that the constrained region is always directly measurable and can be sensed with cameras, which is not always feasible, such as in the case of internal organs or bones.

In LUS, the probe should be placed in the intercostal areas~\cite{Soldati2020ProposalCOVID-19} to avoid the shadowing effect due to the presence of the ribs~\cite{Marini2021LungEssentials}. Detecting the bones and thus placing the probe, mounted on the end-effector of the manipulator, in such a narrow area could become a complex task. In the case of teleoperated robotic ultrasound exams, in particular, standard visual and haptic feedback, fundamental in in-person exams, might not suffice. 
For example, when the robot is in close proximity to the patient, the patient's view captured by a fixed RGB camera might be occluded by the robot. If the camera is mounted on the robot end-effector instead, it might be too close to the patient, reducing the field of view. In such a case, a 3D model, which includes the anatomical features of the patient, such as the rib cage in the case of LUS, and the target pose of the robot, might provide great support. Also, when the probe is in contact with the body, with standard haptic feedback, it is not easy to distinguish if the probe is located on top of a bone or soft tissue, and thus the practitioner should rely only on the ultrasound image.
Another relevant issue is represented by the low repeatability and the high subjectivity of the operations, as the success of the exam depends very much on the practitioner's ability to cope with the anatomical variability of patients.  
\begin{figure*}[t!]
    \centering
    \includegraphics[width=0.85\textwidth]{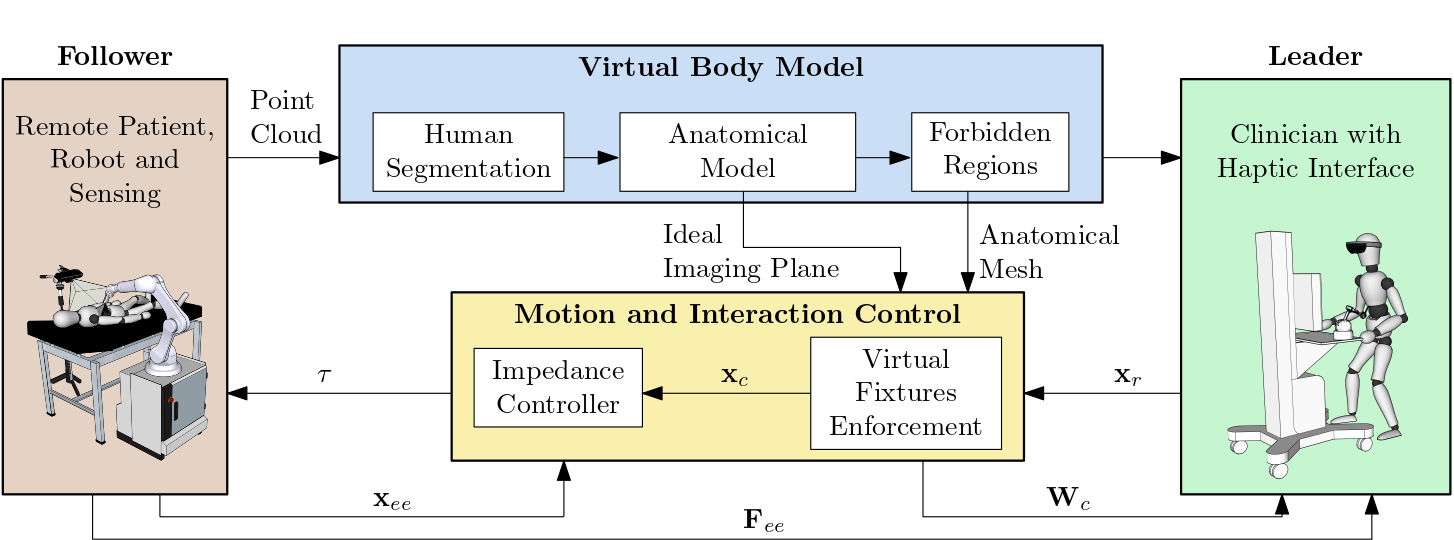}
    \caption{High-level overview of the proposed framework.}
    \label{fig:framework}
    \vspace{-4mm}
\end{figure*}

To overcome these limitations and support the physician during teleoperated procedures, we introduce an anatomy-aware shared control framework for LUS examination. This framework leverages a novel, anatomically accurate 3D model, Skeletal Kinematics Enveloped by a Learned body model (SKEL)~\cite{keller2023skel}, a parametric body model with skin and skeleton meshes, driven by biomechanical pose parameters, not only for reliable 3D visual feedback provided to the clinician but also to define the forbidden regions and the ideal imaging plane of the ultrasound probe.
The main goal of this work lies in assisting the physician during three critical stages of teleoperation:
\begin{enumerate*}[label=\roman*)]
\item during free motion robotic movements, such as reaching a targeted anatomical region, by means of the anatomical 3D visual feedback that facilitates the identification of correct protocol areas;
\item during the alignment phase, by limiting the probe orientation with the ideal computed from the volumetric anatomical model; and
\item during the approach phase, by constraining the probe trajectory to the intercostal spaces through forbidden regions derived from the ribs of the skeletal model.
\end{enumerate*}
The proposed system is classified as Level 1 autonomy (robot assistance) but integrates higher-level features by adapting to different subject anatomies. 
To the best of the authors’ knowledge, this is the first attempt to use accurate anatomical models as a prior to generate real-time, anatomy-aware assistive strategies for teleoperated ultrasound examinations. 
Previous methodologies, in contrast, focused on preoperative data, such as computed tomography~\cite{Li2007SpatialAnatomy}, or ultrasound~\cite{connolly2025touching}, or exclusively on visual data~\cite{ryden2012forbidden,moccia2020vision-based,huang2024robot-assisted}.

We evaluated the framework in a proof-of-concept teleoperated LUS examination procedure. 
We performed the exam in two different experimental conditions. In the first one, 5 naïve subjects performed the exam on the same volunteer, focusing on a subjective evaluation of the technology acceptance, and with an expert user on 6 different volunteers, focusing on a quantitative evaluation of the technology performance.
Overall, we assessed three main different aspects:
\begin{enumerate*}[label=\roman*)]
    \item the precision and accuracy of the SKEL volumetric model to model the patient's body geometry with respect to anthropomorphic measurements;
    \item the capability of the VF to assist the teleoperation in the probe placement within the intercostal areas;
    \item the efficiency of the framework in performing the protocol in terms of duration, compared to a standard teleoperation without assistance.
\end{enumerate*}
The results demonstrate the reliability of the anatomical model, promising results in the usability with naïve subjects and a reduction of the examination duration greater than 20\% with an expert user, highlighting the fundamental role of our assistive and shared control algorithms for an effective remote examination.

\section{METHODOLOGY}
As sketched in \autoref{fig:framework}, the proposed tele-echography architecture couples \begin{enumerate*}[label=\roman*)] \item patient-specific 3D modelling with \item anatomy-aware VF to assist teleoperated robotic LUS. \end{enumerate*}
A coloured point cloud acquired by stereo RGB-D cameras is first segmented with \href{https://github.com/ultralytics/ultralytics}{YOLOv8-seg};
the point cloud is then fitted, via gradient-descent minimisation of the Chamfer distance, to the SKEL volumetric model that simultaneously yields subject-specific skin and skeletal meshes. The skeletal layer localises individual ribs, enabling their axial projection onto the skin surface. The resulting centre lines define elliptical tubes that approximate rib volumes. The generated tubes are encoded as unilateral, mesh-based constraints in a quadratic program (QP) that enforces the VF onto the reference pose of the haptic interface, ensuring the probe slides tangentially over ribs and is naturally funnelled toward intercostal windows. 
Orientation assistance is achieved through conic constraints that bound the probe’s body-fixed axes about an ideal frame whose $z$-axis aligns with the local skin normal and whose $x$-axis follows the rib direction. A second QP shapes the allowable angular velocity, keeping the probe nearly orthogonal to the pleural line while leaving fine alignment to the clinician through shared control. We acknowledge that the real ideal imaging plane is perpendicular to the actual lung surface, which is not always aligned with the body surface. However, its orientation is accessible only when the LUS probe comes into contact with the patient.
The VF-constrained pose is fed to a Cartesian impedance controller that regulates both the motion and interaction of the robot. Contact forces measured at the probe flange close the loop with the 6DoFs haptic interface, while VF activations trigger vibrotactile cues and Graphical User Interface (GUI) overlays of the violated constraint, enhancing situational awareness without masking tissue reaction forces.

\subsection{Patient Body Modelling}
In this section, we introduce the 3D perception pipeline for the generation of the patient-specific model, which includes both the volumetric and the skeletal data of the subject. Although the methodology supports online update of the subject-specific anatomical model from multiple sensing modalities, in the procedure reported here the model was initialised once prior to the examination to allow a focused evaluation of anatomy-aware haptic guidance.

\subsubsection{Anatomical Model using SKEL}
\label{ssec:volumetric_model}

\begin{figure*}[t]
    \centering
    \includegraphics[width=0.95\linewidth]{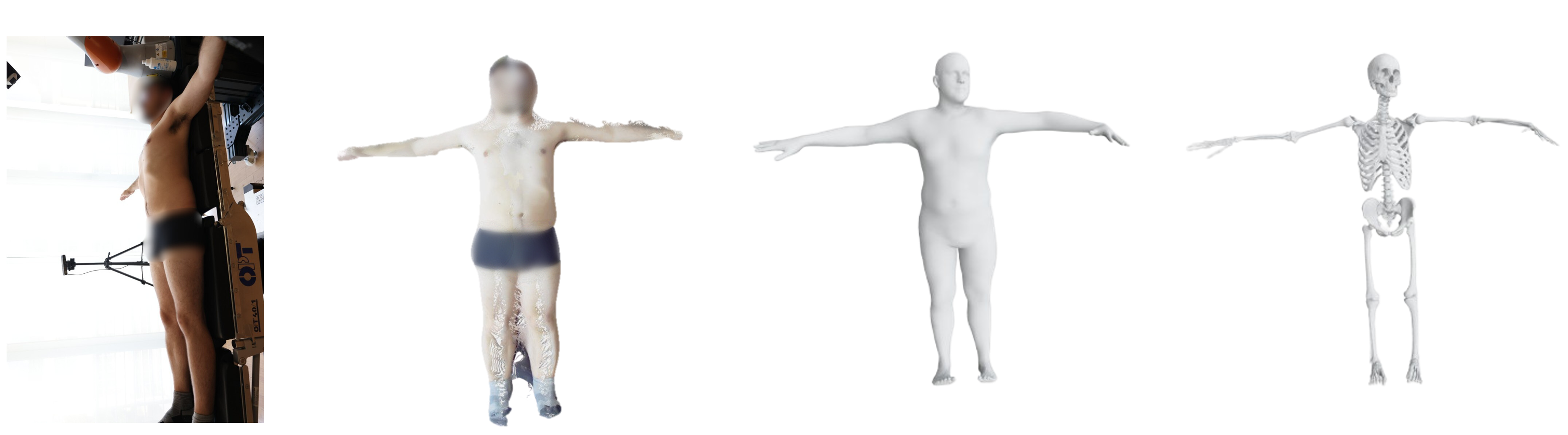}
    
    \caption{Anatomical perception data and models. From left: the RGB-D image of one of the two fixed cameras, the merged point cloud filtered with the YOLO mask, and the reconstructed volumetric and skeletal model. }
    \label{fig:perception_pipeline}
    \vspace{-4mm}
\end{figure*}

The Skeletal Kinematics Enveloped by a Learned body model (SKEL)~\cite{keller2023skel} is a parametric body model with skin and skeleton meshes, driven by biomechanical configuration parameters. With respect to its predecessor, the Skinned Multi-Person Linear (SMPL) model~\cite{loper2015smpl}, it includes an accurate reconstruction of the human skeleton and a biomechanical parametrisation of the model with fewer parameters. 
A SKEL configuration is described by the vector parameter $\boldsymbol{q}_{SKEL} \in \mathbb{R}^{46}$, the shape parameters  $\boldsymbol{\beta}_{SKEL} \in \mathbb{R}^{10}$ and the global position $\boldsymbol{t}_{SKEL} \in \mathbb{R}^{3}$. It comprehends both the volumetric and skeletric model of the human body to capture the anatomical complexity of the human body. For further details on the model accuracy, refer to \cite{keller2023skel}.
To estimate the SKEL model of a subject, the process begins with collecting 3D point clouds of the individual. The body scan is acquired using two calibrated RGB-D cameras mounted on tripods. The method is not limited to two cameras, as incorporating additional cameras with overlapping fields of view can further improve reconstruction accuracy. The acquisition is performed with the subject in T-pose. 
We filter the resulting point clouds using a human segmentation neural network (YOLOv8-seg). Specifically, only the points corresponding to RGB image pixels within the segmentation mask generated by the network are retained. We used the robust ICP algorithm from \href{https://github.com/isl-org/Open3D}{Open3D} 
to merge the two point clouds, compensating for camera calibration inaccuracies.
Subsequently, the position $\boldsymbol{t}_{SKEL}$ is optimised to align SKEL with the 3D scan. Even though SKEL is already in T-pose in its default configuration, $\boldsymbol{q}_{SKEL}$ is refined to adjust the model configuration with respect to the point cloud. Finally, $\boldsymbol{\beta}_{SKEL}$ is estimated to fit the model to the subject’s body shape. The AdamW optimiser is used to minimise the Chamfer distance, which serves as the loss function. The Chamfer distance is computed between the scan points and the vertices of the SKEL skin mesh (volumetric model).
This process estimates $\boldsymbol{t}_{SKEL},\boldsymbol{q}_{SKEL}$ and $\boldsymbol{\beta}_{SKEL}$ in sequence (see~\autoref{fig:perception_pipeline}). 
To ensure a controlled evaluation of the proposed haptic assistance, the volumetric and skeletal anatomical models are initialised once in a T-pose and then used as fixed anatomical models to construct the position- and orientation-based virtual fixtures.
This choice is also motivated by the small magnitude of respiration-related (tidal breathing) and posture-related rib displacements relative to the spatial scale of the virtual fixtures, as well as by the presence of the compliant controller, which naturally accommodates such motion.

\subsubsection{Rib Cage Definition}
With the objective in mind to constrain the interaction of the probe to the skin over the intercostal areas and avoid the ribs, we developed a method to draw, at the skin level, the ribs which lie beneath the skin surface. To define separate anatomical parts of the model, one can just select the triangles of the mesh that belong to the specific anatomical district. Thus, we manually annotate the indexes of superior and inferior boundary vertices of 5 ribs per side (with \href{https://github.com/cnr-isti-vclab/meshlab}{Meshlab}).
We included the ribs from the 1st to the 5th, according to the targeted LUS protocol~\cite{Soldati2020ProposalCOVID-19}.
This needs to be done offline only once, then all the fitted models of the patients will have the same ribs selected.
To draw the rib cage onto the skin of the patient, we used an axial projection along the spine axis. 
We computed the spine axis as the line crossing the manually annotated pelvis and neck vertices of SKEL. Note that since the model is parametric, not only the rib cage but also the spine vertices indices are constant and always referred to the same point of the SKEL model for every possible fitted model.
Hence, we projected every vertex $\boldsymbol{v} \in \mathbb{R}^{3}$ belonging to the inferior and superior border sets towards the skin mesh, obtaining a cylindrical projection of those vertices. The projection direction is defined as the vector starting from the spine axis with minimum distance to $\boldsymbol{v}$ and ending in $\boldsymbol{v}$.
Then, we interpolated the corresponding superior and inferior vertices for each rib with a smooth cubic polynomial. Given these lines, the goal now is to define the 3D forbidden regions of the VF, which, in the case of the rib, looks like a tube. Finally, the 3D mesh of a curved tube with an elliptical cross-section is computed along the central line, retrieved by averaging the superior and inferior cubic polynomials. The ellipse's major axis length is defined to be $\lambda\;\in \mathbb{R}$ times the mean distance between the superior and inferior curves, and the minor axis is ${2}/{3}$ the major axis. While the dimensions of the axes of the ellipse are arbitrary, we would like to highlight that the choice depends on the degree of autonomy we want to retain for the human operator, which depends on his/her experience with teleoperated robotic LUS. In the case of forbidden regions with larger sizes, the reference position is very often constrained with respect to the case of smaller sizes. On the other hand, potential misalignment of the model ribs with the real ones might prevent the operator from placing the probe in the intercostal area. Thus, the choice of the forbidden region size is represented by the tradeoff between the autonomy of the physician and assistance provided by the system.
While the forbidden regions are used to constrain the probe position, we also exploit triangle normals of the volumetric mesh for the probe orientation constraint, which will be explained in the following section.
\subsection{Anatomical Virtual Fixtures}  
Once the ideal orientation and the anatomical forbidden regions are defined, one can enforce VF through a constraint optimisation problem, thereby preventing the ultrasound probe mounted on the end-effector from positioning over the ribs or adopting an improper orientation during the operation. As a result, the system gains robustness with respect to eventual operator imprecise commands due to faulty perception. In this way, the reference coming from the teleoperation is constrained to slide on top of the surface of the virtual fixture until the probe reaches the desired intercostal area. 
In addition, we will prevent the operator from applying high force to the ribs, potentially harming the patient.
Notably, the skeletal model of the body is generated based on the patient's anatomical characteristics, resulting in the forbidden regions constructed on the ribs being tailored to the individual.

\subsubsection{Forbidden Region VF -- for Position}
The position constraint is defined as a unilateral VF, with the forbidden region described by a mesh. The motion control of the robot end-effector is formulated as a quadratic optimisation problem with linear constraints: 
\begin{equation}
    \begin{aligned}
    &\argmin_{\Delta\bs{x}} \;\Vert \Delta \bs{x} - \Delta \bs{ x}_d \Vert_2,\\
    &\text{subject to} \;\;\bs{A} \Delta\bs{x} \geq \bs{b}
\end{aligned}
\label{eq:qp_vf}
\end{equation}
where $\Delta\bs{x} \in \mathbb{R}^3$ and $\Delta\bs{x}_d \in \mathbb{R}^3$ represent, respectively, the actual and the desired incremental position of the end-effector. The linear constraints are defined through the matrices $\bs{A} \in \mathbb{R}^{n \times 3}$ and $\bs{b} \in \mathbb{R}^n$ with $n$ being the number of constraints. 
The rationale behind expressing the problem as a quadratic
minimisation problem is that the forbidden zones can be defined by
multiple walls which are described by hyperplanes with normal
$\boldsymbol{n} \in \mathbb{R}^3$ and its origin $\boldsymbol{p} \in \mathbb{R}^3$. This approach allows for the
imposition of constraints on the movement of the probe tip, ensuring that
only the positive part of the hyperplane is admissible. Given
$d_t, \Delta d \in \mathbb{R}$ the absolute value of the distance between the plane and the probe tip at an instant $t\in \mathbb{R}$, and the change of the signed distance and
$\boldsymbol{x}_{ee}^p~\in~\mathbb{R}^3$ the current end-effector position, respectively, we get:
\begin{equation}
    \begin{aligned}
        &d_{t-1}= \boldsymbol{n}^{\top}(\boldsymbol{x}_{ee}^p-\boldsymbol{p})\\
        &\Delta d = \boldsymbol{n}^{\top}\Delta \bs{x}\\
        &d_t=d_{t-1} + \Delta d \geq 0\\
        &\boldsymbol{n}^{\top}\Delta \bs{x}\geq-\boldsymbol{n}^{\top}(\boldsymbol{x}_{ee}^p-\boldsymbol{p}).
    \end{aligned}
\end{equation}
The linear constraint of the quadratic problem~\eqref{eq:qp_vf} can be
obtained by selecting $\boldsymbol{A} = \boldsymbol{n}^{\top}$ and
$\boldsymbol{b} =
 -\boldsymbol{n}^{\top}(\boldsymbol{x}_{ee}^p-\boldsymbol{p})$. 
In the following subsection, we will explain how the constraint optimisation problem was solved in the case of VF described specifically by meshes.
Li et al.~\cite{li2020anatomical} showed that it is possible to consider
only a local approximation of a mesh to build the optimisation problem. A sphere is defined within which all positions attainable by the robot end-effector $\boldsymbol{x}_{ee}^p$ in a control cycle are contained. The $k$ mesh triangles within this sphere account for the motion constraint. For each triangle $i$, with $i=1$ to $k$, the point at which the distance between the end-effector and the plane defined by the triangle is minimum is determined and defined as the closest point $\mathcal{CP}_i \in \mathbb{R}^3$. If $\mathcal{CP}_i$ is located within the triangle $i$, no alterations are required. Conversely, if $\mathcal{CP}_i$ is positioned outside the triangle, it is necessary to project it onto the edge of the triangle. Subsequently, the planes must be incorporated as active constraints into the minimisation problem. Each triangle plane is defined by a normal $\mathcal{N}_i$ and a point. The plane normals added to the row of the constraint matrix $\boldsymbol{A}$ are determined by the location of $\mathcal{CP}_i$ accounting for these cases:
\begin{enumerate}

\item \textbf{Condition 1}: The $\mathcal{CP}_i$ is in the triangle
  and $\bs{x}_{ee}^p$ is on the positive side of the face normal $\mathcal{N}_i$ (i.e. $\mathcal{N}_i^{T}~(\bs{x}_{ee}^p~-~\mathcal{CP}_i)\ge0$).
 The plane with normal
  $\mathcal{N}_i$ and point $\mathcal{CP}_i$ is added to the constraints.

\item \textbf{Condition 2}: The $\mathcal{CP}_i$ is on the edge of the
  triangle and the local surface described by the considered triangle
  and the adjacent triangle which shares $\mathcal{CP}_i$ is
  convex. The plane with normal $\boldsymbol{x}_{ee}^p-\mathcal{CP}_i$ and point $\mathcal{CP}_i$ is
  added to the constraints.
  
\item \textbf{Condition 3}: The $\mathcal{CP}_i$ is on the edge of the
  triangle, $\bs{x}_{ee}^p$ is on the positive side of the face
  normal $ \mathcal{N}_i$ and the local surface described by the
  considered triangle and the adjacent triangle which shares
  $\mathcal{CP}_i$ is concave. The plane with normal $\mathcal{N}_i$ and point $\mathcal{CP}_i$ is
  added to the constraints.
\end{enumerate}

\subsubsection{Conic Constraints VF -- for Orientation}
Ultrasound image quality highly depends on the probe orientation, since specular reflection of the ultrasound beam is maximised when the probe is positioned orthogonally. 
Specifically in LUS, optimal acquisition requires maintaining the probe in the transverse orientation, which means parallel to the ribs and perpendicular to the pleural line, to avoid the rib shadowing. Achieving and maintaining such an orientation during teleoperation can be particularly challenging. To address this, we leveraged the volumetric body model to derive the ideal orientation $\boldsymbol{R}^c \in \mathbb{R}^{3 \times 3}$ of the ultrasound probe. We defined the coordinate frame of the probe such that the $z$-axis points outward from the sensor, while the $x$-axis extends in the image lateral direction. The volumetric model provides valuable information about the curvature of the surface being approached. Accordingly, during each control cycle, we compute the average normal vector of the neighborhood of $\boldsymbol{x}_{ee}^p$ (for practical use, approx. 30 triangles represent a good neighbourhood). 
The obtained vector is used to define the $z$-axis of $\boldsymbol{R}^c$. The $x$-axis of $\boldsymbol{R}^c$ is determined by projecting the ribs’ extension axis onto the plane orthogonal to the $z$-axis, and the $y$-axis is then orthogonal to the previous axes. Note that, since the model might not be sufficiently accurate and the lung surface is not always parallel to the skin, $\boldsymbol{R}^c$ is in general suboptimal. For this reason, we leave the task of finding the optimal orientation to the physician by looking at the LUS data, in a shared control fashion.
One possibility to enforce a constraint on the orientation is defining conical regions centred on $\bs{R}^c$ that restrict the allowable motion of the rotation matrix axis. We would like to enforce
\begin{equation}
    \boldsymbol{e}_i\trsp \boldsymbol{R} \boldsymbol{e}_i - \cos(\theta_i) \geq 0, i \in \{1,2,3\}
\label{eq:orientation_constraint}
\end{equation}
where $\boldsymbol{R} \in \mathbb{R}^{3\times3}$ is the reference orientation rotation matrix, $\bs{e}_i = [\delta_{i1}\ \delta_{i2}\ \delta_{i3}]^{\top}$, with $\delta_{ij}$ being the Kronecker delta and $i, j \in \{1, 2, 3\}$, and $\theta_i \in(0, \pi / 2)$ determining the size of the cone.
The transformed axis $\bs{R}\bs{e}_i$ is constrained in the conic region with axis $\bs{e}_i$ and angle $2\theta_i$. We would like to define the conic constraints in a different frame ${c}$. To do so, we can rewrite \eqref{eq:orientation_constraint} as
\begin{equation}
    \boldsymbol{e}_i^w\trsp\boldsymbol{R}\trsp \boldsymbol{e}_{i}^c - \cos(\theta_i) \geq 0, i \in \{1,2,3\}
\label{eq:transformed_orientation_constraint}
\end{equation}
where $\boldsymbol{e}_i^w $ corresponds to the $i$ axis of the world frame and $\boldsymbol{e}_{1:3}^c:=\boldsymbol{R}^c\boldsymbol{e}^w_{1:3}$, i.e. the $i$ axis of the $\mathrm{c}$ frame expressed in the world frame.
We can enforce the three conic constraints defined by $\theta_i, i \in {1,2,3}$ in a similar way we did it for the position constraints by means of a QP that constrains the reference angular velocity $\bs{u}_r = \log(\bs{R}_{old}\trsp \bs{R}) \in \mathbb{R}^3$, where $\bs{R}_{old}\in \mathbb{R}^{3\times3}$ is the constrained orientation at the previous time step and $\log(\cdot): SO(3) \rightarrow \mathfrak{s o}(3)$ is the logarithmic map. The optimisation is formulated as:
\begin{equation}
\begin{aligned}
    & \argmin_{\boldsymbol{u}} \;\;\Vert \bs{u} - \bs{u}_r \Vert_2,
    \;\;\text{subject to} \\
    -\bs{e}_i^{c}\trsp \bs{R} [\bs{e}_i^w]_{\times} \bs{u} &\geq - \gamma(\bs{e}_i^w\trsp\bs{R}\trsp \bs{e}_{i}^{c} - \cos(\theta_i)), \forall{i} \in \{1,2,3\}
\end{aligned}
\label{eq:qp_vf_orientation}
\end{equation}
with $\gamma(\cdot)$ being an extended $\mathcal{K}_\infty$ function and $[\cdot]_\times$ the skew-symmetric operator (please refer to \cite{shen2024safe} for an in-depth explanation of the conic constraint).
The idea is that at every control step we update $\boldsymbol{R}_{old}$ by applying an angular velocity $\boldsymbol{u}$ towards $\bs{R}$ that prevents the resulting rotation axes from exiting the constraint cones defined in the $c$ frame.  Since the robot is moving, the nearest triangles will change in time as well as the constraints. To ensure smooth behavior during the robot motion, $\boldsymbol{R}^c$ is converted into a union quaternion and updated using the Slerp function with step $0.1$ which acts as a filter.

\begin{figure}[t!]
    \centering
    \includegraphics[trim={0 4cm 0 7cm},clip,width=0.6\linewidth]{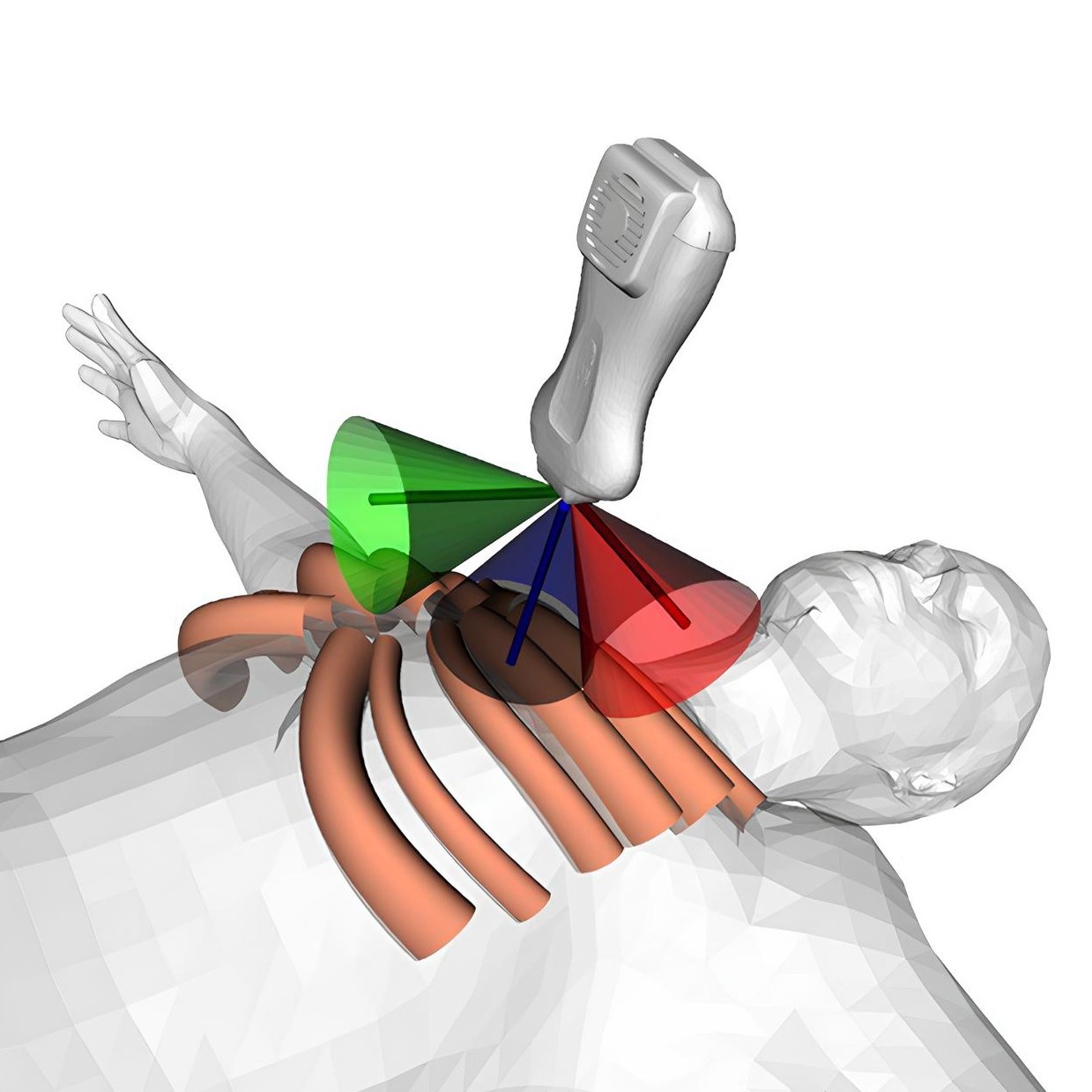}
    \caption{Visual feedback generated by ribs-shaped position VF (salmon), and conic orientation VF. The latter restricts the motion of the axis of the rotation matrix (red for $x$-axis, green for $y$-axis, and blue for $z$-axis).}
    \label{fig:vf_strategy}
    \vspace{-4mm}
\end{figure}

\subsubsection{Visual and Vibrotactile Feedback -- for Position and Orientation} 
During teleoperation, the operator is provided with the view of the two fixed cameras, the ultrasound image, and a GUI. The GUI displays the volumetric model along with the computed rib position constraints, shown from a top-down perspective aligned with the reference frame axes. 
While the position constraints are always visible during the teleoperation, orientation constraints appear on the GUI only when active and are visualised as the corresponding violated constraint cone.
A visualisation of the GUI visual feedback generated by the virtual fixtures strategy for both position and orientation is shown in \autoref{fig:vf_strategy}.
By design, the activation of constraints introduces an error between the reference and the constrained pose; if this error grows too large, controllability may decrease as the constrained pose cannot update due to constraint obstruction. To mitigate this effect, in addition to the end-effector measured interaction force $\boldsymbol{F}_{ee} \in \mathbb{R}^3$, the operator receives an additional feedback wrench $\boldsymbol{W}_c = \begin{bmatrix} \boldsymbol{F}_{c}^T \ \boldsymbol{T}_{c}^T\ \end{bmatrix}^T \in \mathbb{R}^6$ which is generated only when constraints are active. When a position constraint is active, $\boldsymbol{F}_c = [0\ 0\ f_c^z]^T$ is defined as sinusoidal vibration along the z-axis with amplitude $a \in \mathbb{R}$. When an orientation constraint is active, $\boldsymbol{T}_c = b \boldsymbol{\omega}_{err}$, where $b\in\mathbb{R}$ is a scalar gain, and $\boldsymbol{\omega}_{err} \in \mathbb{R}^3$ is the axis-angle error vector between the reference orientation and constrained orientation. The two feedback components are independent and can be active simultaneously. With this approach, the haptic channel of VF acts as a boundary alert, signalling constraint activation rather than continuously steering the user, thus avoiding interference with $\boldsymbol{F}_{ee}$ and preserving unconstrained motion within the feasible region.

\subsection{Robot Interaction Control}
In the context of human-robot interaction and teleoperation, Cartesian impedance control is widely employed due to the possibility to modulate the amount of force exerted from the robot to the environment while tracking Cartesian trajectories~\cite{Sartori2019Tele-echographyScaling, selvaggio2018passive, beber2024passive}. This type of control enables the robot end-effector to behave as a mass-spring-damper system. 
In particular, the amount of force generated depends on the offset between the desired position and the robot actual position and on the impedance parameters $\boldsymbol{D}_d, \boldsymbol{K}_d \in\mathbb{R}^{6 \times 6}$, which are the desired damping and stiffness, respectively. Given $\boldsymbol{W}_{ee} = [\bs{F}_{ee}^T \ \bs{W}_{ee}^T]^T~\in~\mathbb{R}^6$ the interaction wrench between the robot end-effector and the environment, the closed-loop behaviour (no inertia shaping) is described by:
\begin{equation}
    \boldsymbol{\Lambda}\ddot{\Tilde{\boldsymbol{x}}}_{ee} +
    \boldsymbol{D}_d\dot{\Tilde{\boldsymbol{x}}}_{ee} +
    \boldsymbol{K}_d\Tilde{\boldsymbol{x}}_{ee} =
    \boldsymbol{W}_{ee} ,
    \label{eq:cartesian_impedance}
\end{equation}
where
$\Tilde{\boldsymbol{x}}_{ee} = \boldsymbol{x}_r - \boldsymbol{x}_{ee}
\in\mathbb{R}^{6}$ defines the Cartesian position error, $\boldsymbol{x}_r$ the desired position,
$\boldsymbol{\Lambda} \in\mathbb{R}^{6 \times 6}$ the actual inertia. 
Redundancy resolution was used to exploit the redundant degree of freedom. The selected secondary task is joint impedance, with the desired configuration in the middle of the joint angles to prevent joint limits and ensure manipulability.
\section{EXPERIMENTS}

\begin{figure}[t!]
    \centering
    \includegraphics[trim=3.2cm 3.2cm 3.2cm 3.2cm, clip, width=0.9\linewidth]{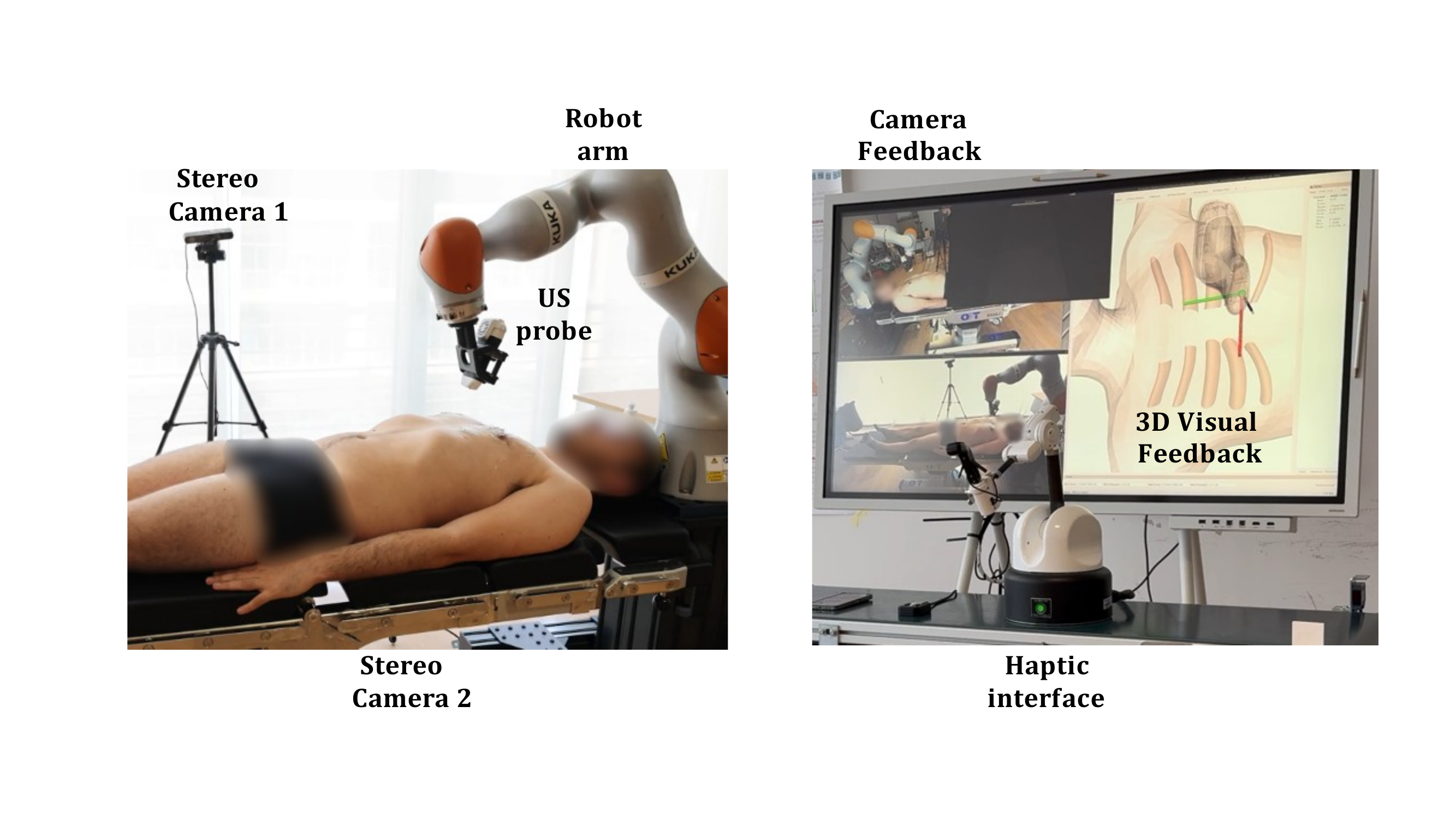}
    \caption{Experimental setup. (left) remote follower side and subject, (right) leader side with haptic interface used by the operator. 
    A video of the experiment is available in the multimedia attachment.}
    \label{fig:setup}
    \vspace{-4mm}
\end{figure}

The proposed telehealth system is divided into two main sites (see~\autoref{fig:setup}). On the patient's side (follower), a collaborative manipulator, capable of estimating the interaction forces, is equipped with an ultrasound probe mounted on its end-effector to perform the physical examination.
The manipulator is teleoperated remotely by an operator by means of a haptic interface (leader), which sends the desired pose to the follower and renders its physical interaction with the environment, i.e., the patient's body, and the vibrotactile feedback through force feedback. Two RGB-D cameras monitor the follower side and broadcast data to the physician's GUI at the leader side. 
The haptic device is a Haption Desktop 6D, which ensures 6 DoFs and force/torque feedback at $1\,$kHz. The manipulator is a KUKA LBR iiwa 14, torque-controlled at $1\,$kHz with a Cartesian impedance controller with diagonal stiffness $K_{i,i}=1000$\;N/m for the translation and $K_{i,i}=30$\;N/rad in orientation and double-diagonalisation damping factor $\xi=1.0$. $\boldsymbol{F}_{ee}$ was constrained such that its $x$, $y$, and $z$ components did not exceed $10$\;N for safety reasons. 
The \mbox{RGB-D} sensors are Zed 2 stereo cameras mounted on tripods for the 3D model generation and 2D visual feedback.
A Clarius PAL HD3 ultrasound probe (linear + phased array) was attached to the robot flange with a 3D-printed support. We exploited the probe chest preset with a depth of $7\,$cm and a focal point on the pleural line. 
We used \href{https://github.com/coin-or/qpOASES}{qpOASES} 
as the solver of the mesh-based position and orientation constraint QPs. The VF parameters are: $\lambda=3.0$, $\theta_i=0.4$\;rad\;$,i\in\{1,2,3\},$ $\gamma$ is a scaling function of factor $0.01$, while for the vibrotactile feedback we set $a=0.45$, $b=0.2$, and $h=80$\;Hz.
The software architecture was running on ROS2 Humble, Ubuntu 22.04, processor AMD Ryzen 9 5900X (24) @ 3.7 GHz × 12-cores CPU, 32 GB RAM, and NVIDIA RTX 4070 GPU. All the experiments were performed with no additional communication delays between leader and follower.

\subsection{Experimental Protocols}
The data collection and experiments were conducted with the approval of the university’s ethics committee under protocol 2025-003ESA.
\subsubsection{Naïve Operators on one Patient} The first experiment involved 5 healthy volunteers (2 female, 3 male, $27.2 \pm 0.8\,$years old). 
The volunteers were naïve with respect to the scope of the experiment and declared to not have relevant previous experience with teleoperation, haptic devices and even video gaming. The volunteers had to play the role of the physician and perform a LUS on a healthy patient ($1,75\,$m, $76\,$kg, 25 years old)  both with our method and with a teleoperation baseline with 2D visual feedback and force feedback.
The volunteers were asked to execute a teleoperated 4-point LUS examination on the anterior part of the chest, with the 4 points selected among the 12 points of the LUS procedure presented by Soldati et al.~\cite{Soldati2020ProposalCOVID-19}. 
Each subject was given 2\;minutes per point to visualise the pleural line on the US data, for a total of 8 minutes total duration.
Before data acquisition, the participants were trained with the fundamentals of LUS through a theoretical explanation (10\;minutes) and a familiarisation phase with the system with both methodologies (10\;minutes each). We tried to average the learning effects by alternating the sequence of the execution of the baseline and our method.

\begin{figure}[!t]
    \centering
    \includegraphics[width=0.75\columnwidth]{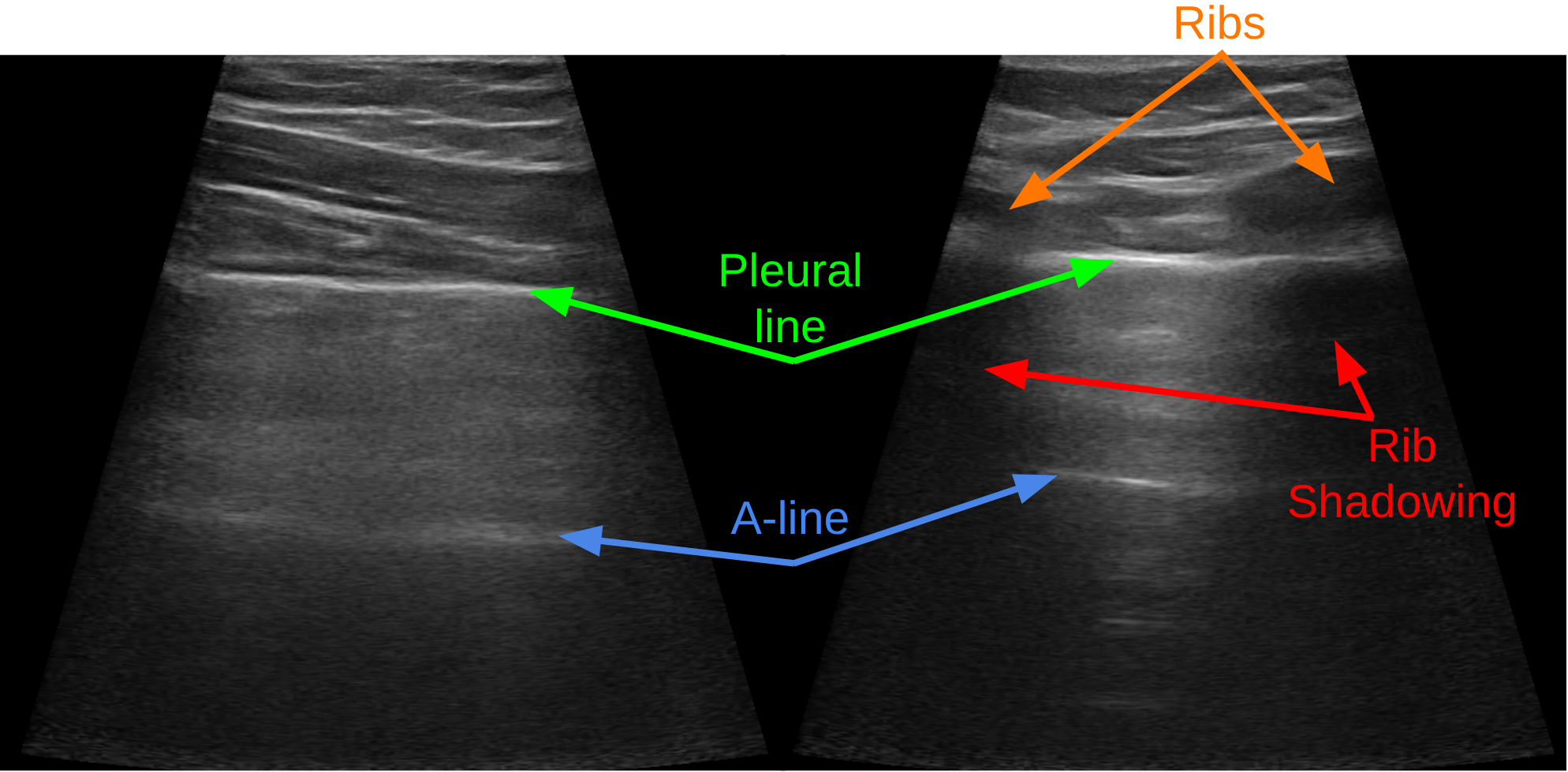}
    \caption{Examples of lung ultrasound images sampled from the US probe at the end-effector of the robot: (left) full pleural line, (right) partial pleural line due to rib shadowing.}
    \label{fig:pleural}
    \vspace{-4mm}
\end{figure}

The goal of this experiment was to evaluate if, with naïve subjects, there are significant improvements (both objective and perceived) of the proposed method with respect to the baseline. For this reason, after each attempt, the subject was required to fill in a NASA TLX questionnaire.  
In light of the challenges associated with the objective assessment of LUS acquisitions due to the lack of image-based metrics to evaluate the quality of LUS images, we compared the methodologies based on the duration of the examinations and the number of valid acquisitions. Thus, the duration was measured from the beginning of the 2\;minutes time budget until a LUS valid acquisition. The acquisition was deemed valid if the pleural line was horizontal on the image and visible for $80\%$ approx. of its horizontal extension (see~\autoref{fig:pleural}) and maintained for at least 2\;s in the image and was measured after the end of the experiment, which was always performed for 8\;minutes. After the end of the examination, we evaluated the recorded US
videos and marked the valid acquisitions. In case the acquisition was not valid within the 2\;minutes time budget, we considered it a failure, and we added 2\;minutes to the total time of the examination.

\begin{figure*}[!t]
    \centering
    \includegraphics[width=0.3\linewidth ]{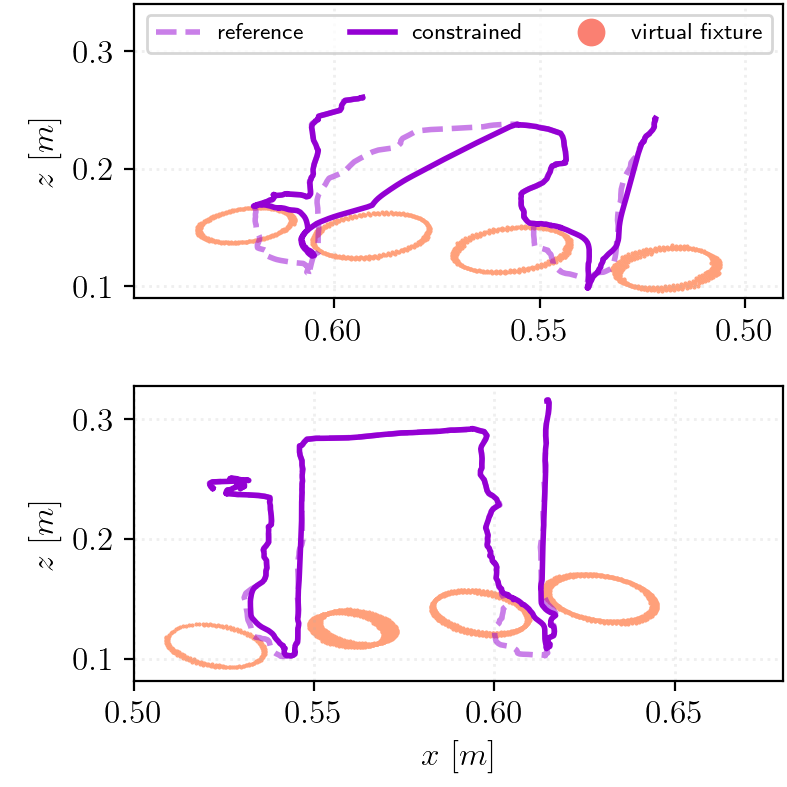}
    \includegraphics[width=0.3\linewidth]{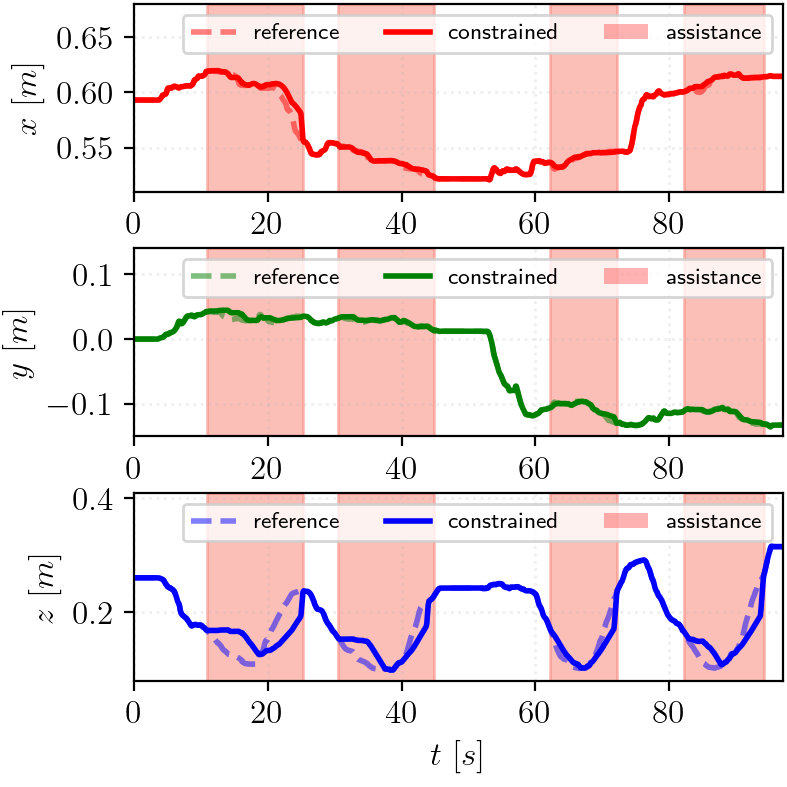}
    \includegraphics[width=0.3\linewidth]{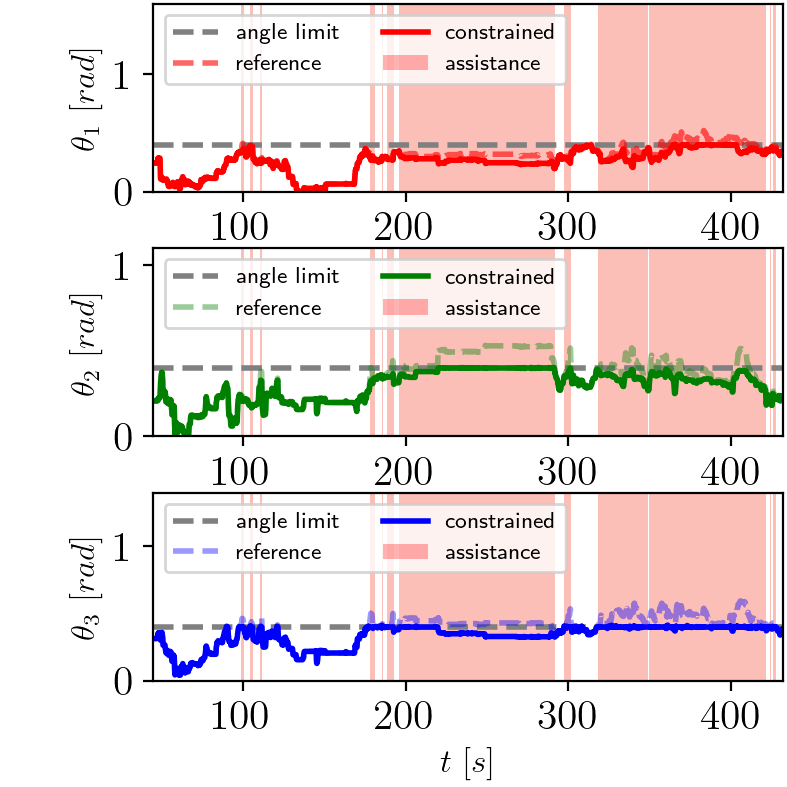}
    \caption{Plots of the experiment described in Section~\ref{sssec:exp_fixtures}. (left) 2D paths, (middle) position trajectories, and (right) orientation trajectories.}
    \label{fig:exp_fixtures}
    \vspace{-4mm}
\end{figure*}

\subsubsection{Expert Operator on 6 Patients}
A second experiment was performed by an expert operator on 6 healthy volunteers acting as patients (6 males, $1.75\pm0.08\,$m, $81 \pm 14\,$kg, $27.3 \pm 1.8\,$years old).

The goals of this experiment were to evaluate the robustness of the systems with respect to different anatomies of patients and to evaluate the improvement of performance with an expert user (trained for 1 hour per 5 days). Only male subjects were included in this study to ensure consistency with the anthropometric measurements with the selected male SKEL model. Although a female model is available, it was not considered in this evaluation, as it would have required a separate anatomical evaluation. While the performance was evaluated through exam duration, as in the previous experiment, we aimed to evaluate the system's robustness with respect to different subjects through the accuracy of the anthropometrical measurements and the mean failure rate across subjects. Since the accuracy of the skeleton reconstruction is strictly related to the volumetric model prediction, we measured the accuracy of the volumetric fitting algorithm, highlighting the capability of SKEL to adapt to different body scans. We considered 3 anthropometric measurements of the chest: 
\begin{enumerate*}
    \item nipple distance (ND),
    \item right nipple to belly button distance (NRBB), and
    \item left nipple to belly button distance (NLBB).
\end{enumerate*}
We acquired these measurements with a measuring tape on the volunteers' bodies and compared them with SKEL volumetric reconstructions extracted with the utility \href{https://github.com/DavidBoja/SMPL-Anthropometry}{SMPL-Anthropometry}.

\subsection{Results and Discussion}

\subsubsection{Validation of the Virtual Fixtures Strategy} \label{sssec:exp_fixtures}
Before evaluating the performance of the system with different subjects, we would like to highlight the effect of the VF on the operator reference, using the data of one LUS executed with our methodology. In~\autoref{fig:exp_fixtures} we show the reference trajectory recorded by the haptic interface and the constrained trajectory, which enforces the VF, which is sent as the desired pose of the robotic arm end-effector. It is possible to observe the activation of the VF in the presence of a mismatch between the reference and the constrained trajectory, both in position (especially in the $z$-axis) and orientation. It is possible to see that, through the VF strategy, the probe slides smoothly within the ribs with an orientation which is close to the ideal one, thus avoiding the undesired rib areas. We would like to point out that the presence of the anatomical 3D visual and the vibrotactile feedback due to the VF constraints violation highly contributes to assisting the physician in reaching the intercostal space.

\subsubsection{Naïve Operators on one Patient}
\begin{figure}[t!]
    \centering
    \includegraphics[width=0.49\columnwidth ]{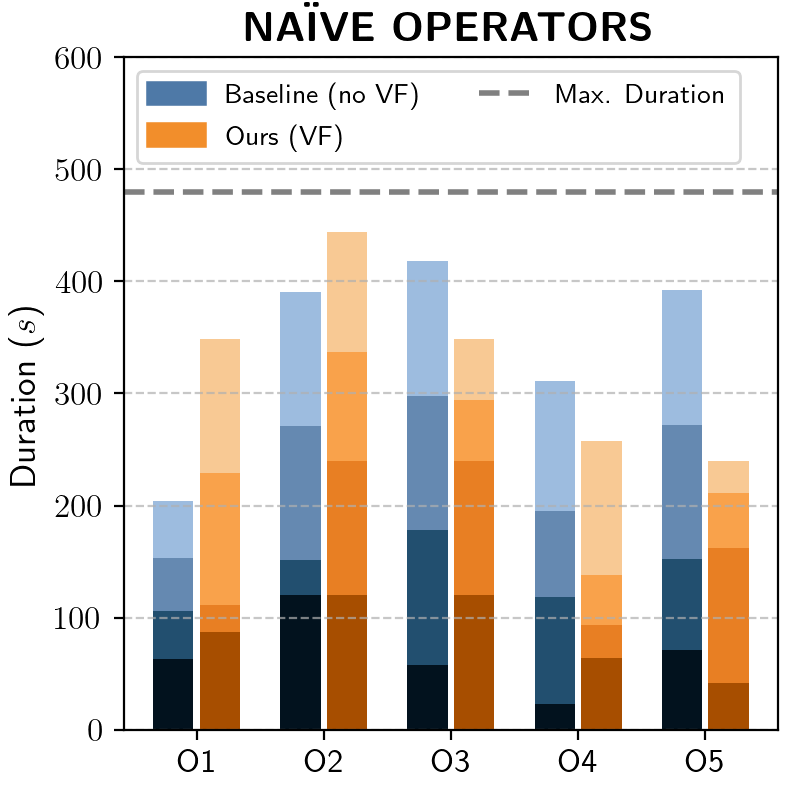}
    \includegraphics[width=0.49\columnwidth]{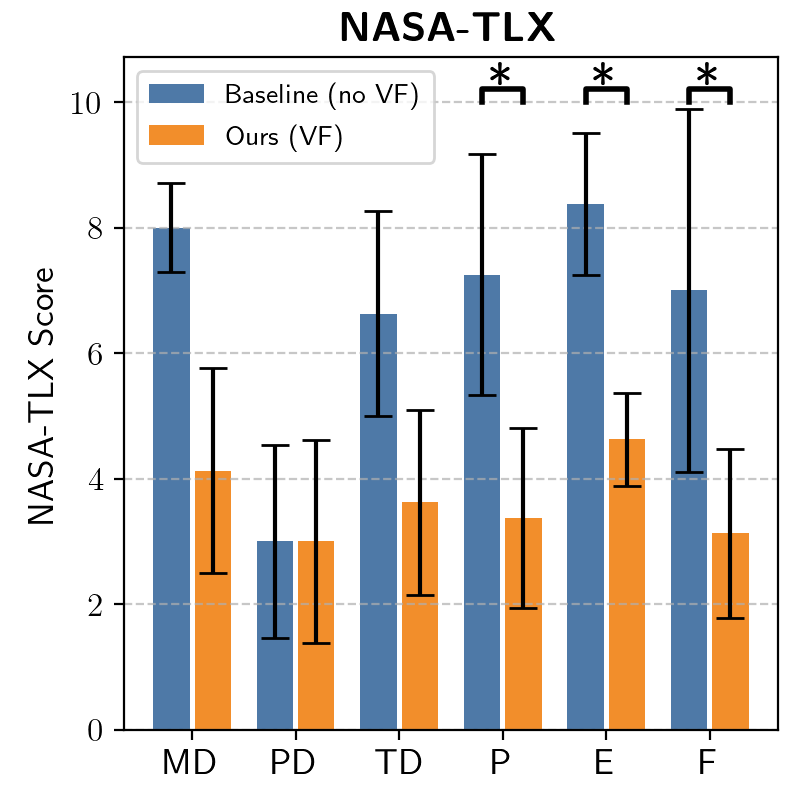}
    \caption{Naïve subjects results in terms of (left) duration of examinations and (right) NASA-TLX subjective evaluation. On the left plot, with the colour gradient, we distinguished the duration related to each point exam. 
    }
    \label{fig:exp_naive}
    \vspace{-4mm}
\end{figure}

The results of the first experiment are displayed in~\autoref{fig:exp_naive}. In general, naïve subjects did not present significant improvement in terms of the duration of the examination ($328\pm82\,$s with our method against $343.2\pm87.5\,$s with the baseline, with $1.4\pm0.55$ failures against $1.6\pm1.52$). Our explanation is that the data of the naïve subjects are affected by the learning curve as the two experiments are performed subsequently, even if we flipped the order of the experiments with different operators. A further analysis with a t-test ($\alpha=0.05$, data normally distributed according to the Shapiro-Wilk test) demonstrated that the presented results did not show statistical significance (paired test p-value $0.783$). However, the perceived experience of the subjects is overall positive, as the NASA-TLX demonstrated that our approach outperforms the baseline, with Performance (p-value $0.006$), Effort (p-value $0.0094$), and Frustration (p-value $0.0278$) reduction being statistically significant ($\alpha=0.05$, data normally distributed according to the Shapiro-Wilk test).

\subsubsection{Expert Operator on 6 Patients}
\begin{figure}[t!]
    \centering
    \includegraphics[width=0.80\columnwidth ]{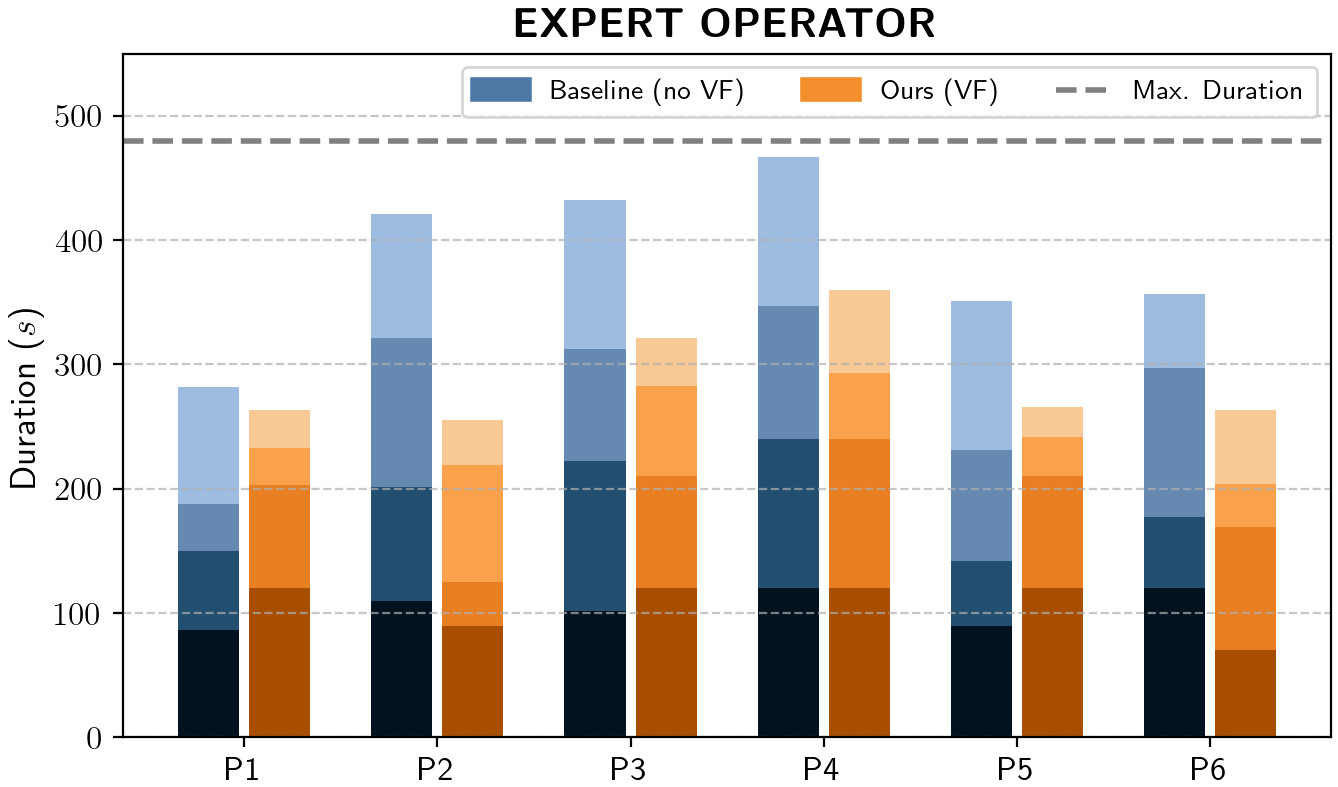}
    \caption{Results of the experiments of the expert operator with 6 patients. While it is possible to highlight that the duration depends much on the anatomy, still the duration with our method results significantly reduced. With the colour gradient, we distinguished the duration of each point exam.}
    \label{fig:exp_expert}
    \vspace{-4mm}
\end{figure}

The second experiment, investigating the performance of the system with a skilled operator and its robustness with respect to the different anatomies, brought much clearer results~\autoref{fig:exp_expert}. First of all, we can recognise a significant improvement in terms of execution time ($293\pm45.7\,$s of our method against $390.6\pm73.9\,$s) and capability to successfully find the pleural line ($1\pm0.7$ faliures against $1.4\pm1.14$). On average, the time reduction of the examination is approx. $23\%$, which is, however, an underestimation, as failures, which are more frequent in the baseline, add only 2 minutes to the total time, whereas in realistic settings the physician has to continue the examination until finding the pleural line in all the inspection points.
The statistical analysis ($\alpha=0.05$, data normally distributed according to the Shapiro-Wilk test) also proved the significance of the results (p-value $0.015$). 
We want to note that, while the data of the first experiment encompass a single patient, here the expert operator had to cope with patients with different anatomies, which requires our system to be robust with respect to anatomical variability. To further evaluate this aspect, we also report the anatomical reconstruction errors, which are below $5$ cm, representing reasonable measurements considering the reference was acquired with a measuring tape (as in standard doctor visits): $0.04 \pm 0.02\,$m for ND, $0.03 \pm 0.02\,$m for NRBB, and $0.02 \pm 0.02\,$m for NLBB. The body-fitting procedure took approximately 2 minutes per patient, although further code optimisation could speed up the process.

\section{CONCLUSIONS}
In this work, we addressed the problem of patient specificity in the context of robotised LUS. Through SKEL, a model that embeds the pose, the shape, and the skeleton of the human body, we were able to provide anatomical references for ultrasound examinations. We then implemented an automatic generation of VF from anatomy, both for position and orientation, that constrains the physical interaction of the probe to the intercostal areas targeted by LUS. A preliminary study validated the framework both quantitatively and subjectively, and results demonstrate potential in assisting clinicians in complex tele-ecography tasks.
Future work includes (i) evaluating online updates of the anatomical model using real-time visual data (e.g., point clouds or skeleton tracking) and (ii) augmenting this refinement with contact-based sensing (e.g., force measurements), to enable viscoelastic body representations as in~\cite{beber2024passive}.
Together, these extensions would support dynamic updates of the anatomical model and overcome the
limitations of purely vision-based approaches.
We also plan to conduct further studies dealing with communication latencies as in~\cite{Kastritsi2024PassiveDelays} and involving multiple clinicians and a larger pool of subjects.

\balance
\bibliographystyle{IEEEtran}
\bibliography{bibliography,virtual_fixture,robot_lung_us}

\end{document}